\documentclass[runningheads]{llncs}

\usepackage[final,year=2026,ID=12464]{eccv}

\usepackage{wrapfig}

\usepackage{eccvabbrv}

\usepackage{graphicx}
\usepackage{booktabs}

\usepackage[accsupp]{axessibility}  

\usepackage{hyperref}
\usepackage{wrapfig}
\usepackage{orcidlink}
\usepackage{xspace}
\usepackage{multirow}
\usepackage{microtype}
\usepackage{bm}
\usepackage[table]{xcolor}
\definecolor{colorRegular}{HTML}{F2F2F2} 
\definecolor{colorInter}{HTML}{EAF4FF}   
\definecolor{colorIntra}{HTML}{F0FFF0}   
\definecolor{colorOther}{HTML}{FFF5EE}   
\definecolor{colorMAP}{HTML}{F3E5F5}     

\begin{document}

\newcommand{\ours}{\texttt{MAP}\xspace}
\title{\ours: Map-Level Attention Processing for Hallucination Mitigation in Large Vision-Language Models} 

\titlerunning{\ours: Map-Level Attention Processing}

\author{Chenxi Li\inst{1}\thanks{\parbox[t]{\linewidth}{
Equal contribution. $^\dagger$ Corresponding author.\\Submitted to ECCV 2026.}} \and Yichen Guo\inst{1}$^\star$  \and Benfang Qian\inst{1} \and Jinhao You\inst{1} \and Kai Tang\inst{1} \and Yaosong Du\inst{1} \and Zonghao Zhang\inst{1} \and Xiande Huang\inst{1}$^\dagger$}


\authorrunning{C.~Li et al.}

\institute{DAIL Tech\\
\email{\{chenxili, yichenguo, benfangqian, jinhaoyou, kai, yaosongdu, zonghaosevenzhang, xdhuang\}@dail.email}}


\maketitle

\begin{abstract}
Hallucinations in Large Vision-Language Models (LVLMs) undermine their reliability in real-world applications, motivating researchers to explore decoding strategies that effectively alleviate this issue. Previous works primarily focus on mitigating hallucinations by refining intermediate hidden states within either inter- or intra-layer scopes, overlooking potentially faithful information beyond these single-dimensional representation spaces. In this paper, we investigate an underexplored map-level perspective by interpreting all hidden states as a two-dimensional semantic map, and empirically demonstrate that faithful information is not localized in a single dimension but is widely dispersed across this 2D map through a systematic logit-lens analysis. Building upon this observation, we propose \textbf{M}ap-Level \textbf{A}ttention \textbf{P}rocessing (\ours), a training-free decoding method that alleviates visual hallucinations from a holistic map-level perspective. Specifically, we design a Layer-Wise Criss-Cross Attention module to progressively refine intermediate hidden states by gathering faithful information from the constructed 2D semantic map. Additionally, a Global-Local Logit Fusion strategy fuses hierarchical content at the logit level to further enhance the robustness of the model's output. Extensive experiments demonstrate that our approach effectively mitigates hallucinations in both closed- and open-ended generation tasks across diverse LVLM architectures.
\end{abstract}

\section{Introduction}
\label{sec:intro}

Recent advances in multimodal fusion and alignment techniques \cite{sun_eva-clip_2023, li_align_2021} have driven the rapid progress in Large Vision-Language Models (LVLMs). Through large-scale pretraining on Internet image-text datasets, LVLMs acquire a broad spectrum of semantic knowledge and demonstrate strong cross-modal reasoning and understanding capabilities \cite{li_survey_2025, shu_large_2025}. These capabilities make LVLMs highly promising for complex cross-modal tasks, such as Visual Question Answering (VQA) \cite{huynh_visual_2025} and Visual Grounding (VG) \cite{yan_vigor_2025}, where accurate understanding and response to multimodal inputs are crucial.

\begin{figure}[h]
\centering
\includegraphics[width=0.7\columnwidth]{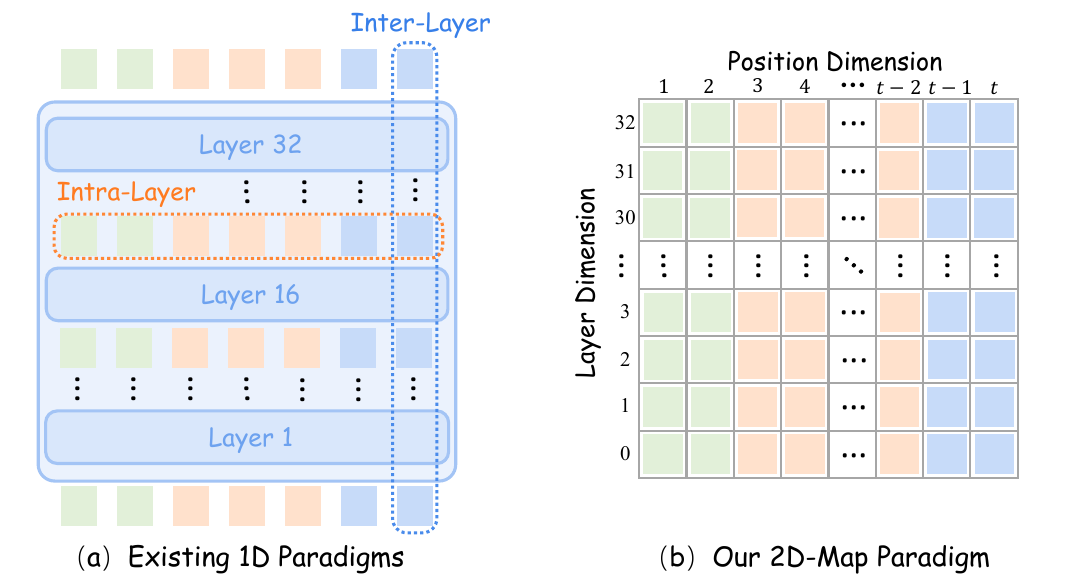}
\vspace{-4.0pt}
\caption{Comparison of existing 1D and our proposed 2D-Map paradigms. (a) Existing methods primarily focus on utilizing information from localized representations within either the inter-layer or intra-layer dimensions. (b) In contrast, our approach reinterprets the entire set of hidden states as a 2D semantic map, thereby enabling a more holistic integration of information across both the position and layer dimensions.}
\label{fig:motivation1}
\vspace{-4.0pt}
\end{figure}

However, visual hallucinations remain a significant obstacle to the broader development of LVLMs \cite{zhao2025aligning, wang_evaluation_2023, rawte_survey_2023}. Specifically, LVLMs often generate responses that are linguistically and grammatically accurate but inconsistent with visual facts, e.g., producing content that includes incorrect objects, spatial relations, or quantities that deviate from the given image. While such hallucinations may be tolerable in casual conversations, they pose serious challenges in domains that require precise and reliable outputs, such as medical imaging, industrial manufacturing, and autonomous driving \cite{gu_medvh_2024, fan_hallucination_2024, jain_semantic_2024}.

Several efforts to mitigate hallucinations primarily focus on knowledge-level enhancements, such as Supervised Fine-Tuning (SFT) \cite{jiang_hallucination_2024, liu_mitigating_2024}, Reinforcement Learning from Human Feedback (RLHF) \cite{zhao_beyond_2024, jing_fgaif_2025, kim_exploiting_2025}, or Retrieval-Augmented Generation (RAG) \cite{qu_alleviating_2025, sarwar_filterrag_2025}. While effective, the high computational cost of these fine-tuning-based methods limits their scalability in real-world applications. More recently, there has been growing interest in training-free approaches that intervene solely during the inference process of LVLMs, including contrastive decoding \cite{wang_mitigating_2024, chuang_dola_2023, leng_mitigating_2023}, guided decoding \cite{wang_mllm_2025}, anchor token reallocation \cite{huang_opera_2024}, visual attention amplification \cite{sarkar_mitigating_2025}, and layer-wise consistency \cite{wang_damo_2024, tang_mitigating_2025, yu2025hallurnn}. 

While these methods have demonstrated promising results, they remain limited to \textbf{single-dimensional} paradigms, as illustrated in Figure~\ref{fig:motivation1}. Specifically, inter-layer methods, such as contrastive decoding and layer-wise consistency, compare and align information across different decoder layers (e.g., hidden states and logits). In contrast, intra-layer methods, such as anchor token reallocation and visual attention amplification, focus on refining token representations within a single decoder layer. However, we pose a critical and underexplored question: \textit{Is there helpful information beyond inter- or intra-layer regions that contributes to hallucination mitigation?} To explore this, we analyze the distribution of factual signals across the full set of hidden states. Specifically, we conduct our analysis using a vanilla LLaVA-1.5 model \cite{liu_visual_2023} and 3,000 images from the random split of the POPE benchmark \cite{li_evaluating_2023}. Each image is paired with a fixed instruction prompt: ``Please describe this image in detail''. Inspired by the logit lens, we project all hidden states across layers and token positions into vocabulary distributions using the language head, and analyze their confidences for in-image and hallucinated object words.

\begin{figure}[t]
\centering
\includegraphics[width=1.0\columnwidth]{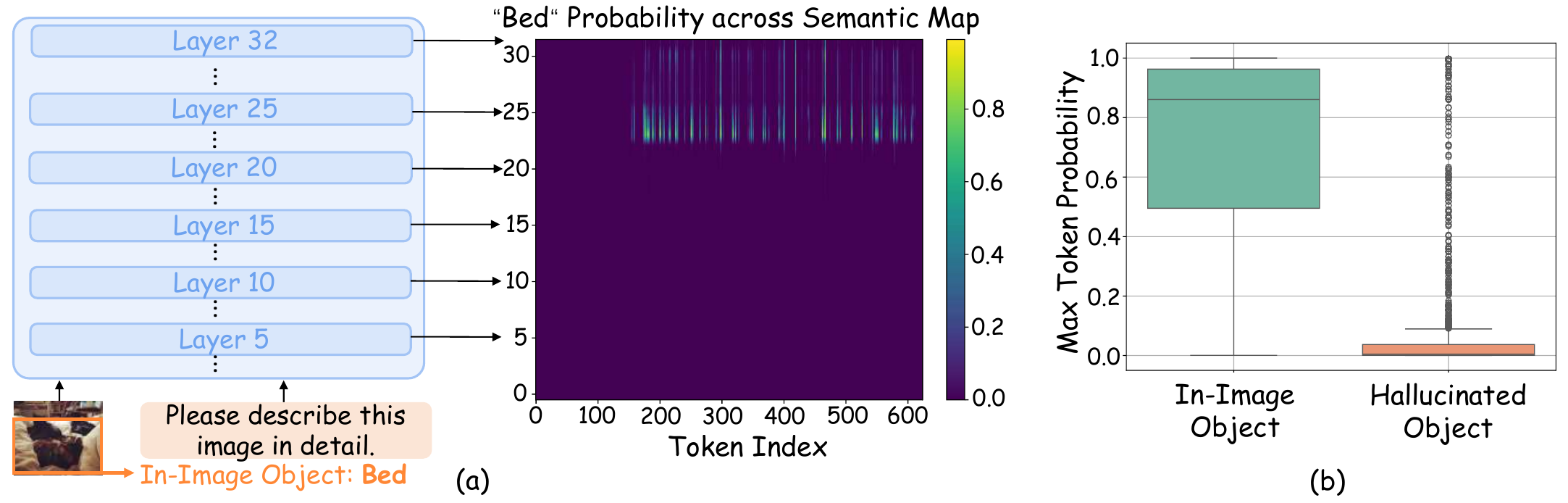}
\caption{(a) A qualitative example showing that faithful semantics are distributed across intermediate hidden states. (b) Statistical comparison showing that in-image objects consistently receive higher token probabilities than hallucinated objects.}
\label{fig:motivation2}
\vspace{-4.0pt}
\end{figure}

A qualitative example is shown in Figure~\ref{fig:motivation2} (a). Given an image containing the object ``bed'', we observe that some tokens within a region (highlighted in red) exhibit high probabilities for the word ``bed''. This indicates that certain informative tokens, including those from intermediate layers and token positions, can still retain factual information relevant to the factual object semantics. Figure~\ref{fig:motivation2} (b) further presents statistical evidence based on 3,000 images. For each image, we compute vocabulary distributions from all intermediate hidden states and record the maximum probability for both in-image and hallucinated object words. The results show a clear, distinct pattern: in-image objects consistently receive higher maximum probabilities, while hallucinated objects are assigned uniformly low confidence. These findings suggest that factual signals are more broadly distributed throughout the latent space, offering potential cues to suppress hallucinations during decoding.

Motivated by this observation, we propose a novel decoding framework, \ours, which interprets the entire set of hidden states as a unified two-dimensional semantic map structured along both the layer and position dimensions. It consists of three key components: Map-Level Operations, Layer-wise Criss-Cross Attention, and Global-Local Logit Fusion. Map-Level Operations aggregate informative tokens dispersed across the 2D semantic map, enabling the model to capture signals beyond conventional inter- or intra-layer boundaries. To refine token representations, we employ a Layer-wise Criss-Cross Attention module that simultaneously captures inter-layer and intra-layer dependencies, facilitating better semantic alignment across layers and positions. We also incorporate a Global-Local Logit Fusion strategy to aggregate hierarchical content, thereby enhancing the robustness of the model’s output by balancing fine-grained local evidence with broader contextual information. Extensive experiments on both closed- and open-ended generation benchmarks demonstrate that \ours effectively reduces visual hallucinations across various LVLM architectures. The contributions are summarized as follows:
\begin{itemize}
\item We present a novel perspective to mitigate hallucinations in LVLMs by treating the inference process as a 2D semantic map, and observe that tokens outside traditional inter- or intra-layer regions can also contribute to hallucination mitigation. 
\item We propose a novel decoding method, \ours, which reduces hallucinations by employing the Map-Level Operations, Layer-wise Criss-Cross Attention module, and Global-Local Logit Fusion technique.
\item We evaluate \ours through extensive experiments on three benchmarks, demonstrating its effectiveness in mitigating hallucinations across both open- and closed-ended scenarios.  
\end{itemize}

\section{Related Work}

\subsection{Hallucinations in LVLMs}
Hallucinations are a widely studied problem in Large Language Models (LLMs), commonly defined as the generation of text that is factually inaccurate or unsupported by the input context \cite{rawte_survey_2023,sriramanan_llm-check_2024,huang_survey_2025}. In the realm of Large Vision-Language Models (LVLMs), hallucinations refer to outputs that conflict with the visual content, such as mentioning non-existent objects or misrepresenting visual attributes \cite{li_evaluating_2023}. Although the exact causes of hallucinations remain unclear, several factors are believed to contribute to their occurrence. Some studies~\cite{liu2024paying, jiang2025devils, sarkar_mitigating_2025} have revealed that LVLMs are prone to pay more attention to the textual context while paying inadequate attention to visual tokens, which may result in hallucinated generations. Moreover, studies~\cite{zhou2023analyzing,tu2025attention,zhu_ibd_nodate} also point out that co-occurrence bias (i.e., the tendency to predict objects that frequently appear together) and over-reliance on language priors can both contribute to hallucinations. Furthermore, works like ~\cite{huang_opera_2024, kang2025see} have observed issues such as attention sinks and anchor tokens, which may amplify hallucinations.

\subsection{Addressing Hallucinations in LVLMs}
Research on hallucination mitigation has mainly focused on several directions, including fine-tuning with curated datasets \cite{chen_mitigating_2023,wang_mitigating_2024}, reinforcement learning \cite{sun_aligning_2023,yu_rlhf-v_2024}, and incorporating external knowledge bases \cite{su_mitigating_2024}. However, these approaches typically require additional data collection or retraining, making them time-consuming and computationally expensive for real-world applications. More recently, generation-time interventions have drawn growing attention. DoLa \cite{chuang_dola_2023}, a contrastive decoding method, compares factual knowledge across shallow and deep layers to encourage faithful generation. VCD \cite{leng_mitigating_2023} extends this by contrasting logits conditioned on visual inputs with different uncertainty levels to enhance robustness. DAMO \cite{wang_damo_2024} emphasizes cross-layer consistency and employs a momentum mechanism to mitigate hallucinations in deep layers. DCLA \cite{tang_mitigating_2025} aggregates hidden states to construct semantic references for consistent decoding. HalluRNN \cite{yu2025hallurnn} applies a modified RNN block across model layers to improve consistency. Other works \cite{huang_opera_2024,gong_damro_2024} identify anchor tokens and reallocate attention weights to mitigate hallucinations. 
\section{Method}

\subsection{Formulation}
Existing LVLMs generally consist of three core components: a vision encoder, a projection module, and a pretrained Large Language Model (LLM). Given an input image $X_v$, the vision encoder $g\left( \cdot \right)$  produces a sequence of visual tokens $Z_v=g\left( X_v \right)$, where $Z_v\in \mathbb{R} ^{L_v\times D_v}$. Here, $L_v$ denotes the number of visual tokens and $D_v$ is the dimensionality of the vision hidden states. These tokens are then projected into the language embedding space as $H_v=P \left( Z_v \right)$, with $H_v\in \mathbb{R} ^{L_v\times D}$, where $D$ is the embedding dimension of the LLM. 

Meanwhile, the system prompt $X_s$ and language query $X_q$ are tokenized using the LLM tokenizer and embedded to obtain system tokens $H_s\in \mathbb{R} ^{L_s\times D}$
and query tokens $H_q\in \mathbb{R} ^{L_q\times D}$, respectively. The visual and textual tokens are then concatenated to form the LLM input: $H_0=\left\{ h_{1,0},...,h_{t,0} \right\}$, where the total sequence length is $L_s+L_v+L_q$. This sequence is passed through the LLM's transformer layers. Let $H_j=\left\{ h_{1,j},...,h_{t,j} \right\} $ represent the hidden states at the $j$-th transformer layer. The hidden state at the final layer $n$ and position $t$ is used by the language head $\phi(\cdot)$ to predict the next token:
\begin{equation}
p\left( x_{t+1}|x_{<t+1} \right) =\text{softmax}\left( \phi \left( h_{t,n} \right) \right) 
\end{equation}

Instead of directly using the final token $h_{t,n}$ to predict the next word, we refine this token representation at the map level, thus effectively reducing visual hallucinations. Formally, the semantic map at the $j$-th decoding layer for a sequence of length $t$ is denoted as:
\begin{equation}
\mathcal{H}_j =
\left[
\begin{array}{cccc}
    h_{1,j} & h_{2,j} & \cdots & h_{t,j} \\
    h_{1,j-1} & h_{2,j-1} & \cdots & h_{t,j-1} \\
    \vdots & \vdots & \ddots & \vdots \\
    h_{1,1} & h_{2,1} & \cdots & h_{t,1}
\end{array}
\right]
\label{eq:semantic-map}
\end{equation}
Where $h_{u,v}\in \mathbb{R} ^D$ is the hidden state of the $u$-th token at the $v$-th decoding layer. Notably, the size of the semantic map $\mathcal{H}_j$ grows progressively with the depth of the decoding layers.

\subsection{The Proposed MAP} 

\begin{figure*}[!t]
\centering
\includegraphics[width=1.0\textwidth]{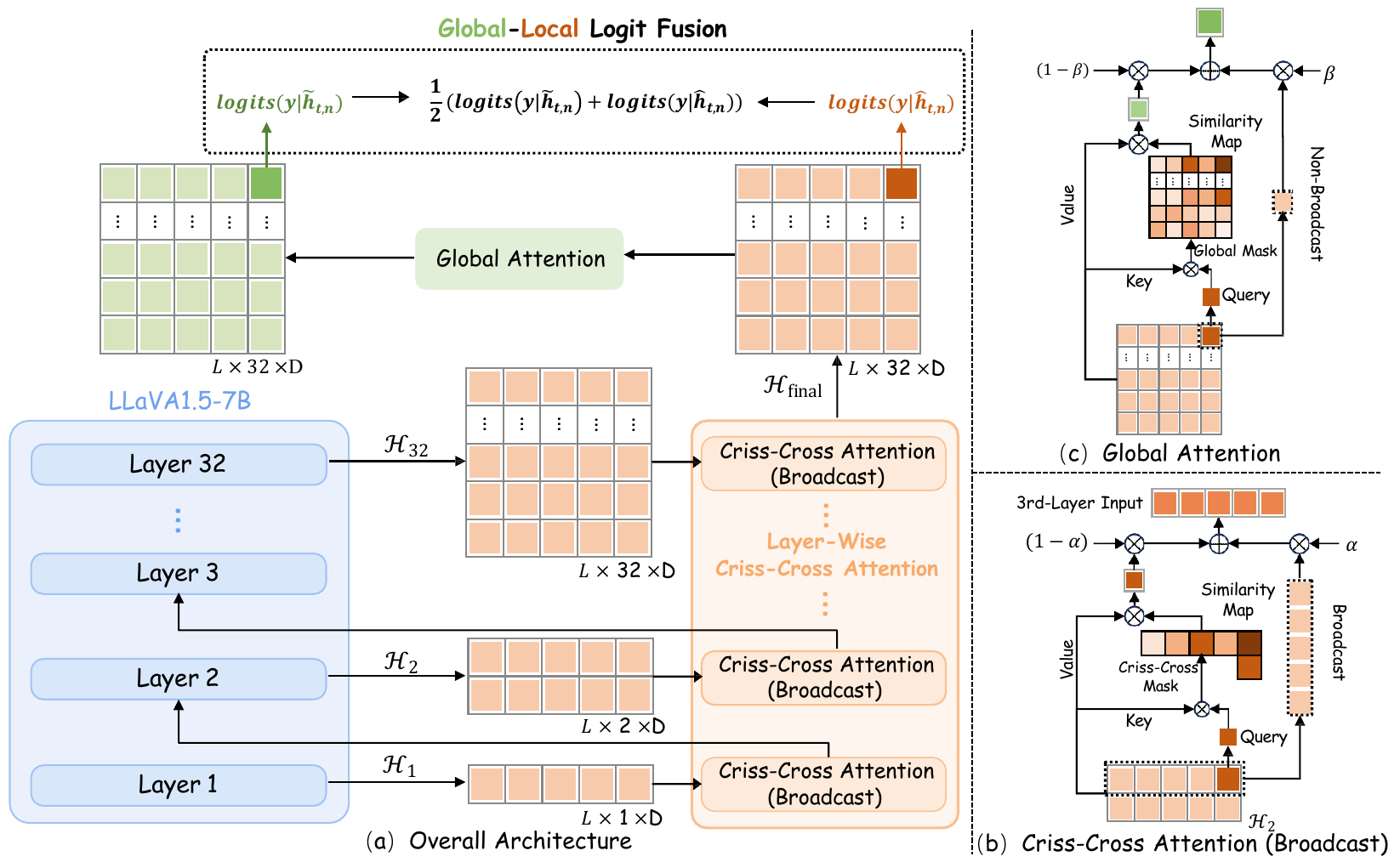}
\caption{Overall architecture of \ours applied to LLaVA1.5-7B. Specifically, \ours introduces a Layer-Wise Criss-Cross Attention module to refine token representations by aggregating faithful information from a 2D semantic map constructed after each decoding layer. Additionally, a Global-Local Logit Fusion module is integrated to fuse hierarchical content at logit-level to produce reliable predictions.}
\label{fig:method}
\end{figure*}

\noindent \textbf{Layer-Wise Criss-Cross Attention.} Using the 2D semantic maps constructed at each layer, we introduce Layer-Wise Criss-Cross Attention to mitigate visual hallucinations during the decoding process, as illustrated in Figure \ref{fig:method}. Specifically, given the semantic map \( \mathcal{H}_j \) at the decoding layer \( j \), we first define the criss-cross semantic neighborhood \( M_c(t,j, \mathcal{H}_j) \), a set of neighboring tokens that share the same row or column index with the anchor token $h_{t,j}$, which the detailed definition is shown in Eq.~\ref{eq:criss-cross}. Then, we introduce a map-level operation function \(\mathcal{F}(\cdot)\) to aggregate all tokens from the defined semantic neighborhood, resulting in an aggregated representation $\mathcal{F}(h_{t,j}, M_c(t,j,\mathcal{H}_j) )$. Subsequently, we update all tokens in layer \( j \) by residualizing the aggregated representation to the original token sequence via a broadcast mechanism:
\begin{equation}
\hat{h}_{u,j} = (1 - \alpha) \cdot \mathcal{F}(h_{t,j}, M_c(t,j,\mathcal{H}_j)) + \alpha \cdot h_{u,j},\ u \in \{1,\dots,t\}
\label{eq:layer-wise criss-cross}
\end{equation}
Where \(\alpha \in [0,1]\) is a trade-off hyperparameter balancing the original and aggregated representations, and the definition of \(\mathcal{F}(\cdot)\) is detailed in  Eq.~\ref{eq:map_function}. As a result, the refined sequence $\left[\hat{h}_{1,j}, \dots, \hat{h}_{t,j} \right]$ integrating factual signals from the semantic neighborhood is then passed to the next transformer layer, modulating the inference flow. 

Notably, our method only takes the last token as the query, naturally reducing computational overhead compared with vanilla attention that takes entire sequence as the query input. By performing Eq.~\eqref{eq:layer-wise criss-cross} in a layer-wise manner, the model could progressively access all token information across the 2D semantic map, achieving a global receptive field with lower computational costs. In practice, to ensure coherent reasoning, we introduce an $\ell_{\text{start}}$ hyperparameter to control from which layer the refinement begins.

\begin{table*}[h]
\centering
\caption{Complementary strengths of local and global logits on the MME benchmark, motivating our Global-Local Logit Fusion, which aggregates both advantages.}
\resizebox{\textwidth}{!}{
\begin{tabular}{lccccccccccc}
\toprule
& Celebrity & Color & Artwork & Count & Landmark & Existence & Posters & OCR & Position & Scene & Total \\
\midrule

Local Logits & 137.4 & 165.0 & 115.0 & \textbf{\textcolor[HTML]{C55A11}{165.0}} & \textbf{\textcolor[HTML]{C55A11}{165.3}} & 190.0 & \textbf{\textcolor[HTML]{C55A11}{144.6}} & 140.0 & 138.3 & \textbf{\textcolor[HTML]{C55A11}{159.3}} & 1520.3 \\
Global Logits & \textbf{\textcolor[HTML]{85BC5F}{139.1}} & \textbf{\textcolor[HTML]{85BC5F}{170.0}} & 115.0 & 160.0 & 164.5 & 190.0 & 138.4 & 140.0 & \textbf{\textcolor[HTML]{85BC5F}{143.3}} & 158.5 & 1519.4 \\
Fused Logits & 138.2 & \textbf{170.0} & \textbf{115.8} & \textbf{165.0} & \textbf{165.3} & 190.0 & 142.5 & 140.0 & \textbf{143.3} & \textbf{159.3} & \textbf{1529.3} \\
\bottomrule
\end{tabular}
}
\label{tab:global_local}
\end{table*}

\vspace{0.1cm}
\noindent \textbf{Global-Local Logit Fusion.} Instead of using refined \(\hat{h}_{t,n}\), we further enhance the final prediction by applying global attention over the final semantic map. Given the final map \(\mathcal{H}_{final}\), we compute a globally enhanced token \(\tilde{h}_{t,n}\) using the global neighborhood \(M_g(t,n, \mathcal{H}_{final})\) anchored at \(\hat{h}_{t,n}\):
\begin{equation}
\tilde{h}_{t,n} = (1 - \beta) \cdot \mathcal{F}(\hat{h}_{t,n}, M_g(t,n,\mathcal{H}_{final})) + \beta \cdot \hat{h}_{t,n}
\end{equation}
where $n$ is the final decoding layer index and $\beta \in [0,1]$ controls the fusion weight. The \(M_g(t,n, \mathcal{H}_{final})\) is detailed in Eq.~\eqref{eq:global}. Then, we fuse the logits obtained from the local token \(\hat{h}_{t,n}\) and the global token $\tilde{h}_{t,n}$ to further  aggregate hierarchical contents. As shown in Table~\ref{tab:global_local}, we empirically demonstrate that the two token logits exhibit complementary strengths across different tasks. For example, the local token performs better on \textit{Count} and \textit{Posters}, while global token excels in \textit{Position} and \textit{Color}. Motivated by this, we fuse the logits from both representations for the final prediction:

\begin{equation}
\left\{
\begin{aligned}
&\text{logit}_{\text{final}} = \frac{1}{2} \cdot \left( \phi(\tilde{h}_{t,n}) + \phi(\hat{h}_{t,n}) \right) \\
&p(x_{t+1} \mid x_{<t+1}) = \text{softmax} \left( \text{logit}_{\text{final}} \right)
\end{aligned}
\right.
\end{equation}

\subsection{Map-Level Operation}\label{sec:map-level-operation}
We detail the map-level operations that form the foundation of both Layer-Wise Criss-Cross Attention and the Global-Local Logit Fusion introduced earlier. Given a 2D semantic map \(\mathcal{H}_j \), we define a map-level operation function $\mathcal{F}(\cdot)$ that computes an aggregated representation by gathering information from a structured semantic neighborhood $M(t,j, \mathcal{H}_j)$ anchored at the token $h_{t,j}$. Formally, $\mathcal{F}(\cdot)$ is represented as:
\begin{equation}
\mathcal{F}(h_{t,j}, M(t,j,\mathcal{H}_j)) = \sum_{h_{u,v} \in M} f(h_{u,v}, h_{t,j}) \cdot h_{u,v}
\label{eq:map_function}
\end{equation}
where \(f(\cdot, \cdot)\) is a cosine similarity function that measures the relevance between a neighboring token $h_{u,v}$ and the query token $h_{t,j}$. Similarity scores are normalized via softmax to ensure proper weighting. By varying the definition of the semantic neighborhood $M(\cdot)$, different map-level operations can be instantiated. Next, we describe the designs of global and criss-cross semantic neighborhoods.

\vspace{0.1cm}
\noindent \textbf{Global Semantic Neighborhood.} Global attention aggregates contextual information from the entire semantic map. In this setting, the global semantic neighborhood \(M_g(t,j,\mathcal{H}_j)\) includes all hidden states in \(\mathcal{H}_j\) except the anchor token itself: 
\begin{equation}
M_{g}(t,j,\mathcal{H}_j) = \{ h_{u,v} \in \mathcal{H}_j \mid (u,v) \neq (t,j) \}
\label{eq:global}
\end{equation}

\noindent \textbf{Criss-Cross Semantic Neighborhood.} Criss-cross attention captures factual information along the row and column of the anchor token. Specifically, this criss-cross semantic neighborhood \(M_c(t,j,\mathcal{H}_j)\) consists of all tokens from the same layer $j$ and position $t$, excluding the token itself:
\begin{equation}
M_{c}(t,j) = \{ h_{u,j} \mid u \neq t \} \cup \{ h_{t,v} \mid v \neq j \}
\label{eq:criss-cross}
\end{equation}
\section{Experiment}

\subsection{Experiment Setup}

\noindent \textbf{Datasets.} We evaluate the effectiveness of \ours across three widely used benchmarks that span both closed- and open-ended tasks: POPE~\cite{li_evaluating_2023}, MME~\cite{fu_mme_2024}, and MMHal-Bench~\cite{sun_aligning_2023}. POPE is an object-level hallucination benchmark that comprises three subsets drawn from MSCOCO~\cite{lin_microsoft_2014}, A-OKVQA~\cite{schwenk_-okvqa_2022}, and GQA~\cite{hudson_gqa_2019}. Additionally, MME offers a comprehensive and fine-grained evaluation across ten categories. Finally, we use MMHal-Bench to evaluate the effectiveness of \ours for hallucination mitigation in an open-ended generation setting, in which the responses are automatically assessed by GPT-4o~\cite{hurst2024gpt}.

\vspace{0.1cm}
\noindent \textbf{Models and Baselines.} We compare \ours against recent decoding methods covering both inter-layer and intra-layer approaches, including DAMO~\cite{wang_damo_2024}, DCLA~\cite{tang_mitigating_2025}, SPIN~\cite{sarkar2025mitigating} HGAI~\cite{jiang2025devils}, and VCD~\cite{leng2024mitigating}. Additionally, to assess the generalization of \ours across various LVLM architectures, we select three representative models: LLaVA-1.5-7B~\cite{liu_visual_2023}, mPLUG-Owl2-7B~\cite{ye_mplug-owl2_2023}, and InstructBLIP-7B~\cite{dai2023instructblip}.

\vspace{0.1cm}
\noindent \textbf{Implementation Details.} We have provided all the hyperparameters involved in other methods and \ours, which are available in the Appendix. In addition, we investigated the model's sensitivity to hyperparameters in Section~\ref{exp:ablation}, including $\alpha$, $\beta$, and $\ell_{\text{start}}$, confirming that our hyperparameter is optimal. To ensure reproducibility, we disable model sampling and set the temperature to zero. 

\subsection{Main Experiment Results}

\noindent \textbf{Results on the MME Benchmark.} The results of the MME benchmark are presented in Table~\ref{tab:mme}. As shown, \ours effectively reduced visual hallucinations, reaching 1529.3 on LLaVA 1.5, 1466.4 on mPLUG-Owl 2, and 1302.7 on InstructBLIP. Notably, compared to the regular decoding baseline, \ours yielded significant performance gains on both LLaVA 1.5 and InstructBLIP, with improvements of 37.7 and 31.3 points, respectively. In contrast, intra-layer approaches, such as SPIN, showcased imbalanced performance across different LVLMs, with an increase of 10.3 on LLaVA 1.5 but a decline of 2.5 on mPLUG-Owl2. For inter-layer methods, both DAMO and DCLA demonstrated clear performance improvements across different models, with total scores of 1513.5 and 1520.1 on LLaVA 1.5, respectively. However, \ours effectively outperformed these approaches, demonstrating the effectiveness of our method.

\begin{table}[t]
\centering
\footnotesize
\setlength{\tabcolsep}{3.5pt} 
\renewcommand{\arraystretch}{1.2} 
\caption{Evaluation on MME. 
\colorbox{colorRegular}{Regular}, 
\colorbox{colorOther}{Other}, 
\colorbox{colorInter}{Inter-Layer}, 
\colorbox{colorIntra}{Intra-Layer}, 
\colorbox{colorMAP}{\textbf{Map-Level (Ours)}} methods are shaded accordingly. Best results are in \textbf{bold}.}
\resizebox{\textwidth}{!}{%
\begin{tabular}{l l *{11}{c}} 
\toprule
Model & Decoding & Cele. & Color & Artw. & Count & Land. & OCR & Post. & Exis. & Posi. & Scen. & Total \\
\midrule
\rowcolor{colorRegular} \cellcolor{white} & Vanilla & 135.0 & 165.0 & 119.3 & 160.0 & 162.3 & 125.0 & 140.5 & 190.0 & 138.3 & 156.3 & 1491.6 \\
\rowcolor{colorOther} \cellcolor{white} & VCD \textcolor[HTML]{367DBD}{2024} & \textbf{151.8} & 148.3 & \textbf{125.5} & 148.3 & 165.3 & 130.0 & 131.6 & 190.0 & 118.3 & 150.3 & 1459.4 \\
\rowcolor{colorInter} \cellcolor{white} & DAMO \textcolor[HTML]{367DBD}{2024} & 134.1 & 165.0 & 120.0 & 150.0 & 163.8 & 140.0 & \textbf{144.6} & \textbf{195.0} & 143.3 & 157.8 & 1513.5 \\
\rowcolor{colorInter} \cellcolor{white} & DCLA \textcolor[HTML]{367DBD}{2025} & 132.1 & \textbf{175.0} & 117.3 & 163.3 & 160.5 & 140.0 & 137.4 & 190.0 & \textbf{148.3} & 156.3 & 1520.1 \\
\rowcolor{colorIntra} \cellcolor{white} & SPIN \textcolor[HTML]{367DBD}{2025} & 132.9 & 170.0 & 123.0 & 163.3 & 165.3 & 125.0 & 141.5 & 190.0 & 133.3 & 157.5 & 1501.9 \\
\rowcolor{colorIntra} \cellcolor{white} & HGAI \textcolor[HTML]{367DBD}{2025} & 135.0 & 165.0 & 118.5 & 160.0 & 160.8 & 125.0 & 140.5 & 190.0 & 138.3 & 156.3 & 1489.3 \\
\rowcolor{colorMAP} \cellcolor{white} \multirow{-7.2}{*}{LLaVA1.5} & \textbf{\ours} & 138.2 & 170.0 & 115.8 & \textbf{165.0} & \textbf{165.3} & \textbf{140.0} & 142.5 & 190.0 & 143.3 & \textbf{159.3} & \textbf{1529.3} \\
\midrule
\rowcolor{colorRegular} \cellcolor{white} & Vanilla & 162.9 & 150.0 & \textbf{139.8} & 165.0 & 160.0 & 102.5 & 163.3 & 185.0 & 78.3 & 152.8 & 1459.5 \\
\rowcolor{colorOther} \cellcolor{white} & VCD \textcolor[HTML]{367DBD}{2024} & 162.4 & 150.0 & 135.3 & 145.0 & 147.8 & 95.0 & 149.3 & 175.0 & 83.3 & 151.0 & 1394.0 \\
\rowcolor{colorInter} \cellcolor{white} & DAMO \textcolor[HTML]{367DBD}{2024} & 163.8 & 150.0 & 139.0 & 160.0 & 161.5 & 95.0 & \textbf{163.3} & 185.0 & \textbf{93.3} & 153.8 & 1464.7 \\
\rowcolor{colorInter} \cellcolor{white} & DCLA \textcolor[HTML]{367DBD}{2025} & 163.8 & 155.0 & 139.0 & 165.0 & 159.5 & 102.5 & 162.2 & 185.0 & 78.3 & 153.0 & 1463.4 \\
\rowcolor{colorIntra} \cellcolor{white} & SPIN \textcolor[HTML]{367DBD}{2025} & \textbf{166.2} & \textbf{160.0} & 136.5 & 160.0 & 160.0 & 102.5 & 160.2 & 185.0 & 68.3 & \textbf{158.3} & 1457.0 \\
\rowcolor{colorIntra} \cellcolor{white} & HGAI \textcolor[HTML]{367DBD}{2025} & 162.4 & 150.0 & 136.5 & 165.0 & 160.8 & 102.5 & 157.1 & 185.0 & 86.7 & 153.5 & 1459.4 \\
\rowcolor{colorMAP} \cellcolor{white} \multirow{-7.2}{*}{mPLUG-Owl2} & \textbf{\ours} & 163.5 & 155.0 & 135.8 & \textbf{165.0} & \textbf{162.5} & \textbf{102.5} & 160.2 & \textbf{185.0} & 83.3 & 153.3 & \textbf{1466.4} \\
\midrule
\rowcolor{colorRegular} \cellcolor{white} & Vanilla & 154.7 & 143.3 & 112.5 & 58.3 & 165.8 & 80.0 & 152.7 & 185.0 & 58.3 & \textbf{160.8} & 1271.4 \\
\rowcolor{colorOther} \cellcolor{white} & VCD \textcolor[HTML]{367DBD}{2024} & 127.1 & 140.0 & 107.0 & 68.3 & 133.8 & 63.3 & 117.0 & 145.0 & 63.3 & 149.3 & 1153.2 \\
\rowcolor{colorInter} \cellcolor{white} & DAMO \textcolor[HTML]{367DBD}{2024} & 155.3 & 143.3 & 116.3 & 78.3 & 159.3 & 87.5 & 149.0 & 185.0 & 68.3 & 157.3 & 1299.5 \\
\rowcolor{colorInter} \cellcolor{white} & DCLA \textcolor[HTML]{367DBD}{2025} & 157.6 & 143.3 & \textbf{119.5} & 73.3 & 157.3 & 87.5 & 149.7 & 185.0 & \textbf{68.3} & 158.0 & 1299.6 \\
\rowcolor{colorIntra} \cellcolor{white} & SPIN \textcolor[HTML]{367DBD}{2025} & 152.9 & 143.3 & 114.3 & \textbf{86.7} & 160.3 & 87.5 & 141.2 & 185.0 & 63.3 & 159.8 & 1294.2 \\
\rowcolor{colorIntra} \cellcolor{white} & HGAI \textcolor[HTML]{367DBD}{2025} & 153.8 & 143.3 & 111.0 & 58.3 & 165.0 & 80.0 & \textbf{152.7} & 185.0 & 56.7 & 159.0 & 1264.9 \\
\rowcolor{colorMAP} \cellcolor{white} \multirow{-7.2}{*}{InstructBLIP} & \textbf{\ours} & \textbf{158.5} & \textbf{143.3} & 115.0 & 70.0 & \textbf{165.8} & \textbf{95.0} & 147.3 & \textbf{185.0} & 63.3 & 159.5 & \textbf{1302.7} \\
\bottomrule
\end{tabular}}
\label{tab:mme}
\end{table}

\begin{figure}[t]
\centering
\includegraphics[width=1.0\textwidth]{figs/mmhal.png} 
\vspace{-20pt}
\caption{Evaluation on MMHal-Bench with LLaVA-1.5 under open-ended scenario.}
\label{fig:mmhal}
\end{figure}

\begin{table*}[t]
\centering
\scriptsize
\setlength{\tabcolsep}{2.8pt}
\renewcommand{\arraystretch}{1.4}
\caption{Evaluation on POPE. 
\colorbox{colorRegular}{Regular}, 
\colorbox{colorOther}{Other}, 
\colorbox{colorInter}{Inter-Layer}, 
\colorbox{colorIntra}{Intra-Layer}, 
\colorbox{colorMAP}{\textbf{Map-Level (Ours)}} methods are shaded accordingly. Best results are in \textbf{bold}.}
\resizebox{\textwidth}{!}{%
\begin{tabular}{l l *{18}{c}}
\toprule
\multirow{3}{*}{Model} & \multirow{3}{*}{Decoding} & \multicolumn{6}{c}{MSCOCO} & \multicolumn{6}{c}{A-OKVQA} & \multicolumn{6}{c}{GQA} \\
\cmidrule(lr){3-8} \cmidrule(lr){9-14} \cmidrule(lr){15-20}
& & \multicolumn{2}{c}{Random} & \multicolumn{2}{c}{Popular} & \multicolumn{2}{c}{Adversarial} & \multicolumn{2}{c}{Random} & \multicolumn{2}{c}{Popular} & \multicolumn{2}{c}{Adversarial} & \multicolumn{2}{c}{Random} & \multicolumn{2}{c}{Popular} & \multicolumn{2}{c}{Adversarial} \\
\cmidrule(lr){3-4} \cmidrule(lr){5-6} \cmidrule(lr){7-8} \cmidrule(lr){9-10} \cmidrule(lr){11-12} \cmidrule(lr){13-14} \cmidrule(lr){15-16} \cmidrule(lr){17-18} \cmidrule(lr){19-20}
& & Acc. & F1 & Acc. & F1 & Acc. & F1 & Acc. & F1 & Acc. & F1 & Acc. & F1 & Acc. & F1 & Acc. & F1 & Acc. & F1 \\
\midrule

\rowcolor{colorRegular} \cellcolor{white} & Vanilla & 89.57 & 89.66 & 86.10 & 86.70 & 79.50 & 81.48 & 87.10 & 88.30 & 80.10 & 83.08 & 69.03 & 75.90 & 86.50 & 87.83 & 74.73 & 79.45 & 68.77 & 75.73 \\
\rowcolor{colorOther} \cellcolor{white} & VCD \textcolor[HTML]{367DBD}{2024} & 89.07 & 89.00 & 85.63 & 86.03 & 79.27 & 81.01 & 86.97 & 87.23 & 78.90 & 81.82 & 67.90 & 75.01 & 86.00 & 87.33 & 74.50 & 79.29 & 68.10 & 75.15 \\
\rowcolor{colorInter} \cellcolor{white} & DAMO \textcolor[HTML]{367DBD}{2024} & 90.03 & 90.02 & 86.73 & 87.13 & 80.47 & 82.15 & 87.87 & 88.91 & 81.30 & 83.87 & 70.53 & 76.75 & 87.50 & 88.58 & 76.17 & 80.28 & 70.07 & 76.42 \\
\rowcolor{colorInter} \cellcolor{white} & DCLA \textcolor[HTML]{367DBD}{2025} & 90.03 & 89.99 & 86.73 & 87.10 & 80.70 & 82.28 & 87.93 & 88.98 & 81.13 & 83.78 & 70.23 & 76.60 & 87.90 & 88.94 & 75.20 & 79.68 & 69.13 & 75.91 \\
\rowcolor{colorIntra} \cellcolor{white} & SPIN \textcolor[HTML]{367DBD}{2025} & 89.63 & 89.67 & 86.43 & 86.90 & 79.90 & 81.74 & 87.70 & 88.79 & 80.63 & 83.41 & 69.97 & 76.43 & 87.53 & 88.66 & 75.27 & 79.76 & 68.87 & 76.43 \\
\rowcolor{colorIntra} \cellcolor{white} & HGAI \textcolor[HTML]{367DBD}{2025} & \textbf{90.70} & \textbf{90.77} & \textbf{87.67} & \textbf{88.12} & 81.00 & 82.80 & 87.37 & 88.50 & 80.37 & 83.20 & 70.00 & 76.42 & 87.07 & 88.24 & 74.37 & 79.10 & 69.20 & 75.90 \\
\rowcolor{colorMAP} \cellcolor{white} \multirow{-7}{*}{LLaVA1.5} & \textbf{MAP} & 90.10 & 89.86 & 87.40 & 87.44 & \textbf{81.80} & \textbf{82.82} & \textbf{89.03} & \textbf{89.70} & \textbf{82.70} & \textbf{84.70} & \textbf{72.77} & \textbf{77.84} & \textbf{88.57} & \textbf{89.32} & \textbf{77.90} & \textbf{81.22} & \textbf{72.57} & \textbf{77.70} \\
\midrule

\rowcolor{colorRegular} \cellcolor{white} & Vanilla & 88.27 & 87.56 & 86.37 & 85.82 & 84.07 & 83.82 & 88.27 & 88.23 & 84.73 & 85.21 & 77.57 & 79.67 & 86.56 & \textbf{87.20} & \textbf{80.23} & \textbf{80.67} & 78.43 & \textbf{79.27} \\
\rowcolor{colorOther} \cellcolor{white} & VCD \textcolor[HTML]{367DBD}{2024} & 86.77 & 86.11 & 83.97 & 83.59 & 82.10 & 82.07 & 86.83 & 86.93 & 82.07 & 82.96 & 75.50 & 78.13 & 84.87 & 84.08 & 77.60 & 78.00 & 75.70 & 76.38 \\
\rowcolor{colorInter} \cellcolor{white} & DAMO \textcolor[HTML]{367DBD}{2024} & 88.40 & 87.75 & 86.40 & 85.92 & 83.90 & 83.75 & 88.13 & 88.10 & 84.43 & 84.95 & 77.23 & 79.42 & 87.03 & 86.38 & 79.67 & 80.18 & 77.97 & 78.88 \\
\rowcolor{colorInter} \cellcolor{white} & DCLA \textcolor[HTML]{367DBD}{2025} & 88.53 & 87.99 & 86.03 & 86.05 & 82.67 & 83.25 & 87.87 & 88.23 & 83.77 & 84.82 & 75.33 & 78.63 & \textbf{87.07} & 86.98 & 78.90 & 80.37 & 76.77 & 78.81 \\
\rowcolor{colorIntra} \cellcolor{white} & SPIN \textcolor[HTML]{367DBD}{2025} & 88.47 & 87.71 & 86.63 & 86.03 & 84.37 & 84.04 & 88.13 & 87.96 & 84.43 & 84.77 & 77.60 & 79.46 & 86.10 & 85.40 & 79.80 & 80.11 & 78.13 & 78.81 \\
\rowcolor{colorIntra} \cellcolor{white} & HGAI \textcolor[HTML]{367DBD}{2025} & 88.43 & 87.67 & 86.47 & 85.87 & 84.30 & 83.97 & 88.33 & 88.26 & 84.57 & 85.04 & 77.57 & 79.64 & 86.90 & 86.15 & 79.63 & 80.00 & 78.03 & 78.76 \\
\rowcolor{colorMAP} \cellcolor{white} \multirow{-7}{*}{mPLUG2} & \textbf{MAP} & \textbf{88.80} & \textbf{88.16} & \textbf{86.70} & \textbf{86.25} & \textbf{84.37} & \textbf{84.21} & \textbf{88.33} & \textbf{88.32} & \textbf{84.77} & \textbf{85.27} & \textbf{77.70} & \textbf{79.82} & 86.93 & 86.31 & 80.13 & 80.33 & \textbf{78.57} & 79.10 \\
\midrule

\rowcolor{colorRegular} \cellcolor{white} & Vanilla & 87.13 & 85.72 & 83.50 & 84.09 & 80.77 & 81.93 & 88.60 & 88.34 & 79.47 & 81.85 & 71.37 & 76.38 & 86.93 & 87.34 & 76.47 & \textbf{79.30} & 71.60 & 76.04 \\
\rowcolor{colorOther} \cellcolor{white} & VCD \textcolor[HTML]{367DBD}{2024} & 84.10 & 83.07 & 80.17 & 79.27 & 78.17 & 78.13 & 82.03 & 82.36 & 76.90 & 78.40 & 71.67 & 74.93 & 81.43 & 81.60 & 73.73 & 75.81 & 71.93 & 74.80 \\
\rowcolor{colorInter} \cellcolor{white} & DAMO \textcolor[HTML]{367DBD}{2024} & 88.23 & 87.46 & 84.93 & 84.49 & 82.50 & 82.31 & 87.33 & 87.77 & 79.07 & 81.29 & 71.40 & 76.07 & 87.00 & 87.47 & 75.03 & 78.42 & 72.13 & 76.50 \\
\rowcolor{colorInter} \cellcolor{white} & DCLA \textcolor[HTML]{367DBD}{2025} & \textbf{88.40} & \textbf{87.57} & 83.97 & 82.44 & 82.77 & 81.37 & 88.13 & 88.28 & 79.83 & 81.59 & 72.37 & 76.39 & 87.47 & 87.65 & 75.93 & 78.70 & 73.00 & 76.71 \\
\rowcolor{colorIntra} \cellcolor{white} & SPIN \textcolor[HTML]{367DBD}{2025} & 86.33 & 84.61 & 84.10 & 82.53 & 82.83 & 81.40 & 88.57 & 88.11 & 79.60 & 81.97 & 71.47 & 76.47 & 86.70 & 86.04 & 76.20 & 79.15 & 71.37 & 75.93 \\
\rowcolor{colorIntra} \cellcolor{white} & HGAI \textcolor[HTML]{367DBD}{2025} & 87.30 & 85.87 & 84.73 & 83.49 & 82.80 & 81.79 & 88.57 & 88.35 & 79.30 & 81.77 & 71.33 & 76.40 & 87.30 & 86.99 & 75.57 & 77.66 & \textbf{74.03} & 76.68 \\
\rowcolor{colorMAP} \cellcolor{white} \multirow{-7}{*}{InstBLIP} & \textbf{MAP} & 88.30 & 87.43 & \textbf{85.43} & \textbf{84.82} & \textbf{82.97} & \textbf{82.14} & \textbf{88.60} & \textbf{88.41} & \textbf{80.20} & \textbf{81.85} & \textbf{73.03} & \textbf{76.80} & \textbf{87.57} & \textbf{87.69} & \textbf{76.53} & 79.06 & 73.83 & \textbf{77.20} \\
\bottomrule
\end{tabular}}
\label{tab:pope}
\end{table*}

\vspace{0.1cm}
\noindent \textbf{Results on MMHal-Bench.} MMHal-Bench provides an open-ended generation evaluation for hallucination mitigation across eight sub-tasks, the results of which are presented in Figure~\ref{fig:mmhal}. As illustrated, \ours obtained the highest overall score of 2.4 and showcased balanced performance across most tasks. It can be seen that our method reached 3.8 on \textit{Environment}, improving vanilla LLaVA 1.5 by 0.5. In contrast, although other methods generally gained an improved overall score, they exhibited task-specific instability. For example, DAMO improved by 0.4 on relation but dropped 0.6 on \textit{Counting}; VCD scored 2.9 on \textit{Comparison} but only 1.8 on \textit{Counting}. SPIN performed more consistently but offered limited overall improvement. These results demonstrate the effectiveness of \ours in open-ended generation tasks.

\vspace{0.1cm}
\noindent \textbf{Results on POPE.} Compared to MME, POPE provides a more targeted evaluation of object-level hallucinations. The results under random, popular, and adversarial settings are shown in Table~\ref{tab:pope}. As demonstrated, \ours effectively outperformed most decoding baselines across LVLM architectures, particularly achieving notable performances on LLaVA-1.5 and InstructBLIP. On the challenging GQA adversarial subset, our method outperformed VCD decoding by 4.47\% on LLaVA-1.5 and 2.87\% on mPLUG-Owl2, respectively. In contrast, VCD showcased limited performance across diverse settings. For intra-layer methods, such as HGAI, which reached an accuracy of 90.70\% on LLaVA-1.5 in MSCOCO but only obtained 87.37\% in A-OKVQA, demonstrating its instability. Inter-layer methods, which include DAMO and DCLA, obtained a clear performance improvement compared with VCD. These results highlight the robustness of \ours in mitigating object-level hallucinations across diverse conditions.

\subsection{Ablation Study}
\label{exp:ablation}


\noindent \textbf{Evaluation on Map-Level Operations.} Here, we disable the Global-Local Logit Fusion module and evaluate only the effectiveness of various map-level operations, the results of which are shown in Table~\ref{tab:map_ablation}. We further introduce a local semantic neighborhood for local attention, defined by \(M_l(t,j,\mathcal{H}_j) = \{ h_{u,v} \mid |u - t| \le r, |v - j| \le r, (u,v) \neq (t,j) \}\), where \(r\) denotes the window radius. For non-layer-wise approaches, we apply them only to \(\mathcal{H}_{32}\). As shown in Table~\ref{tab:map_ablation}, it can be observed that all non-layer-wise map operations showed limited improvements with MME score of 1502.5 and 1509.7 for local and global attention, respectively. By performing criss-cross attention in a Layer-Wise  manner, it can increase the score from 1507.9 to 1520.3, outperforming other map operations.

\begin{figure}[t]
\centering
\includegraphics[width=1.0\columnwidth]{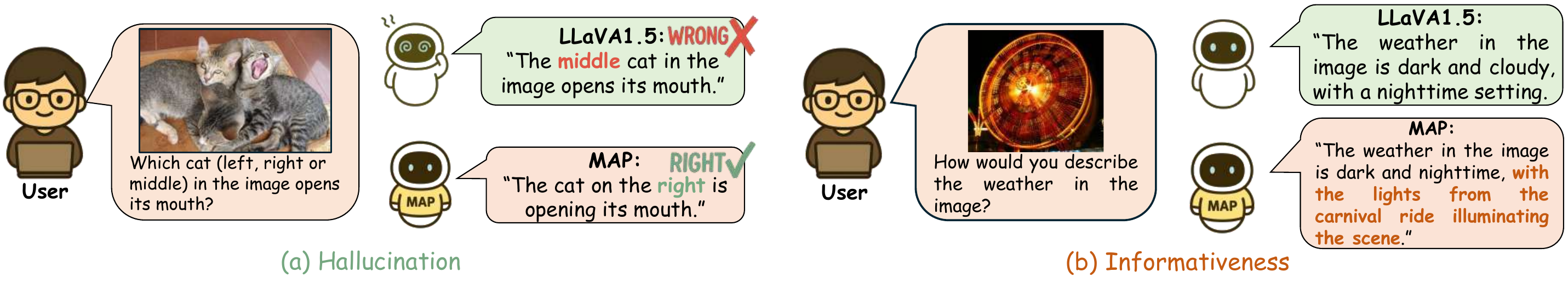}
\caption{Qualitative comparison of \ours in terms of hallucination and informativeness.}
\label{fig:case_study}
\end{figure}

\begin{table}[t]
\centering
\begin{minipage}[t]{0.49\textwidth}
    \centering
    \caption{Performance comparison of various map operations without logit fusion.}
    \label{tab:map_ablation}
    \renewcommand{\arraystretch}{0.9}
    \footnotesize
    \setlength{\tabcolsep}{3pt}
    \begin{tabular}{l c}
    \toprule
    Operation & MME Score \\
    \midrule
    Vanilla & 1491.6 \\
    Global Attention & 1509.7 \\
    Local ($7\times7$) & 1502.4 \\
    Local ($5\times5$) & 1502.5 \\
    Criss-Cross & 1507.9 \\
    \textbf{Layer-Wise Criss-Cross} & \textbf{1520.3} \\
    \bottomrule
    \end{tabular}
\end{minipage}
\hfill 
\begin{minipage}[t]{0.49\textwidth}
    \centering
    \caption{Efficiency analysis. Prefill and decode latencies are reported in ms/token.}
    \label{exp:efficiency}
    \renewcommand{\arraystretch}{0.9}
    \footnotesize
    \setlength{\tabcolsep}{2pt}
    \begin{tabular}{lcccc}
    \toprule
    Method & \begin{tabular}[c]{@{}c@{}}Prefill\\(ms)\end{tabular} & \begin{tabular}[c]{@{}c@{}}Decode\\(ms)\end{tabular} & \begin{tabular}[c]{@{}c@{}}GPU\\(GB)\end{tabular} & TFLOPs \\
    \midrule
    Vanilla & 73.16 & 19.37 & 15.28 & 8.49 \\
    DAMO    & 90.85 & 38.69 & 15.36 & 8.51 \\
    DCLA    & 90.72 & 28.60 & 15.29 & 8.49 \\
    SPIN    & 74.88 & 22.19 & 15.27 & 8.49 \\
    \textbf{\ours} & \textbf{77.33} & \textbf{26.69} & \textbf{15.29} & \textbf{8.49} \\
    \bottomrule
    \end{tabular}
\end{minipage}
\end{table}

\vspace{0.1cm}
\noindent \textbf{Evaluation on Computational Efficiency.} Here, we conduct an efficiency analysis of \ours compared with other methods in the aspect of including prefill latency, decode latency, GPU utilization, and TFLOPs. As shown in Table~\ref{exp:efficiency}, \ours achieved a decode latency of 26.69 ms per token, which is lower than DAMO's 38.69 ms and DCLA's 28.60 ms, demonstrating the decoding efficiency of our method. Additionally, the GPU utilization and TFLOPs of each method are comparable, indicating that these decoding strategies are resource-efficient. Theoretically, compared with the vanilla self-attention approach that treats the entire sequence as the query input, our map-level operation only queries the last token, which reduces the computational complexity from $O(n^2)$ to $O(n)$.

\vspace{0.1cm}
\noindent \textbf{Qualitative Results.} Here, we qualitatively evaluate \ours under an open-ended generation setting, the results of which are presented in Figure~\ref{fig:case_study}. As illustrated, \ours effectively corrected the hallucinated response given by the vanilla model. Additionally, it can also be observed that \ours could further enhance the overall informativeness of the generated response, which added more details about the scene description.

\begin{figure}[t]
  \centering
  \begin{minipage}[t]{0.43\textwidth}
    \vspace{0pt} 
    \captionof{table}{Hyperparameter sensitivity analysis of $\alpha$ and $\ell_{\text{start}}$ on POPE.}
    \label{tab:alpha_layer_sensitivity}
    \vspace{2pt}
    \scalebox{0.78}{ 
    \begin{tabular}{l | c c c c c}
    \toprule
    $\alpha$ & 0.80 & 0.84 & 0.86 & 0.88 & 0.90 \\
    \midrule
    Random      & 90.10 & 90.10 & 90.00 & 90.13 & \textbf{90.16} \\
    Popular     & \textbf{87.40} & 87.33 & 87.33 & 87.30 & 87.27 \\ 
    Adversarial & \textbf{81.80} & 81.50 & 81.53 & 81.50 & 81.47 \\
    \midrule \midrule
    $\ell_{\text{start}}$ & 27-th & 28-th & 29-th & 30-th & 31-th \\
    \midrule
    Random      & 89.17 & 90.00 & 90.10 & \textbf{90.23} & 90.13 \\
    Popular     & 85.70 & 87.00 & 87.40 & \textbf{87.47} & 87.37 \\ 
    Adversarial & 80.17 & 81.47 & \textbf{81.80} & 81.70 & 81.60 \\
    \bottomrule
    \end{tabular}}
  \end{minipage}
  \hfill
  \begin{minipage}[t]{0.55\textwidth}
    \vspace{15pt} 
    \centering
    \includegraphics[width=\textwidth]{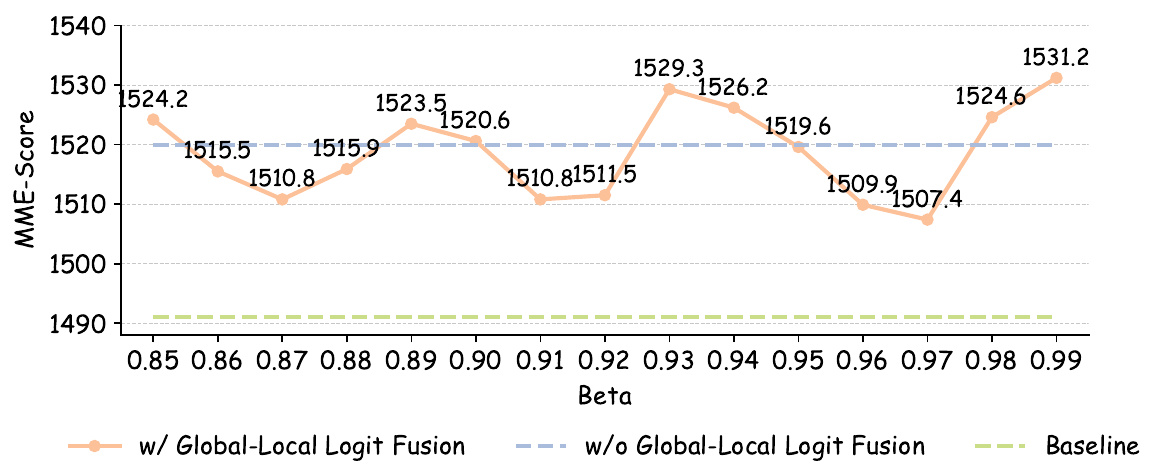}
    \vspace{-12pt} 
    \caption{Hyperparameter sensitivity analysis of $\beta$ on MME.}
    \label{fig5}
  \end{minipage}
\end{figure}


\vspace{0.1cm}
\noindent \textbf{Sensitivity of \bm{$\alpha$} and \bm{ $\ell_{\text{start}}$}.}
We evaluate the sensitivity of the hyperparameter $\alpha$ and $\ell_{\text{start}}$ using LLaVA1.5 on the MSCOCO split of the POPE benchmark. For $\alpha$, we vary its value from 0.80 to 0.90, with results reported in Table~\ref{tab:alpha_layer_sensitivity}. \ours achieved the best average performance when $\alpha=0.80$ and remains stable across the range, with variation within 0.40\%. For $\ell_{\text{start}}$, we test different layers at which to initiate representation refinement, with results shown in Table~\ref{tab:alpha_layer_sensitivity}. The best performance is achieved when $\ell_{\text{start}}=29$.


\vspace{0.1cm}
\noindent \textbf{Sensitivity of \bm{$\beta$}.} We evaluate the impact of $\beta$ in the Global-Local Logit Fusion module by conducting experiments on the MME benchmark with $\beta$ values ranging from 0.85 to 0.99. The total MME scores are shown in Figure~\ref{fig5}. All settings consistently outperform the regular decoding baseline, though slight fluctuations are observed compared to the variant without logit fusion. Notably, the best performance is achieved at $\beta = 0.99$, demonstrating the effectiveness of the Global-Local Logit Fusion module.

\begin{wraptable}{r}{0.45\textwidth} 
  \centering
  \vspace{-30pt} 
  \caption{Generalization of \ours to more advanced LVLMs on the MME.}
  \label{exp:stronger_lvlm}
  \footnotesize
  \setlength{\tabcolsep}{2pt} 
  \scalebox{0.8}{
  \begin{tabular}{l c c c}
    \toprule
    Method & QwenVL2.5 & InternVL2.5 & InternVL3 \\
    \midrule
    Vanilla & 1704.2 & 1697.2 & 1729.6 \\
    DAMO    & 1699.9 & 1709.9 & 1725.8 \\
    DCLA    & 1704.0 & 1694.6 & 1726.4 \\
    SPIN    & 1686.3 & 1685.2 & 1722.8 \\
    \textbf{MAP} & \textbf{1712.0} & \textbf{1704.3} & \textbf{1737.0} \\
    \bottomrule
  \end{tabular}}
  \vspace{-10pt} 
\end{wraptable}

\vspace{0.1cm}
\noindent \textbf{Evaluation on Stronger LVLMs.} To evaluate the effectiveness of \ours on more advanced LVLMs, we further conduct experiments on three models: Qwen2.5-VL-7B \cite{bai2025qwen2}, InternVL2.5-8B \cite{chen2024expanding}, and InternVL3-14B \cite{zhu2025internvl3}. The MME total scores presented in Table~\ref{exp:stronger_lvlm} demonstrate that \ours effectively enhances the models' performance, with the total scores increasing from 1704.2 to 1712.0 on Qwen2.5-VL-7B and from 1697.2 to 1704.3 on InternVL2.5-8B, respectively. Moreover, when applied to the large-scale 14B model, our method also shows robustness, with a 7.4-point improvement on InternVL3-14B, highlighting the generalization of \ours across different LVLMs.


\section{Conclusion}
In this work, we present a novel paradigm that views all hidden states as a 2D semantic map, expanding the single-dimensional perspectives focused by existing methods. Through a logit lens analysis, we reveal that intermediate hidden states beyond intra- or inter-layer regions also showcase a certain level of factual semantics that could contribute to hallucination mitigation. Motivated by this, we propose \ours, a decoding method for mitigating hallucinations in LVLMs. \ours introduces Layer-Wise Criss-Cross Attention to progressively refine token representations during decoding, and Global-Local Logit Fusion to enhance outputs at the logit level. Extensive experiments validate the effectiveness of \ours and verify the map-level paradigm in mitigating hallucinations.

\bibliographystyle{splncs04}
\bibliography{main}

\title{Supplementary Material} 
\appendix

\clearpage
\section{Additional Results for Preliminary Experiments}

\begin{figure}[t]
\centering
\includegraphics[width=1.0\columnwidth]{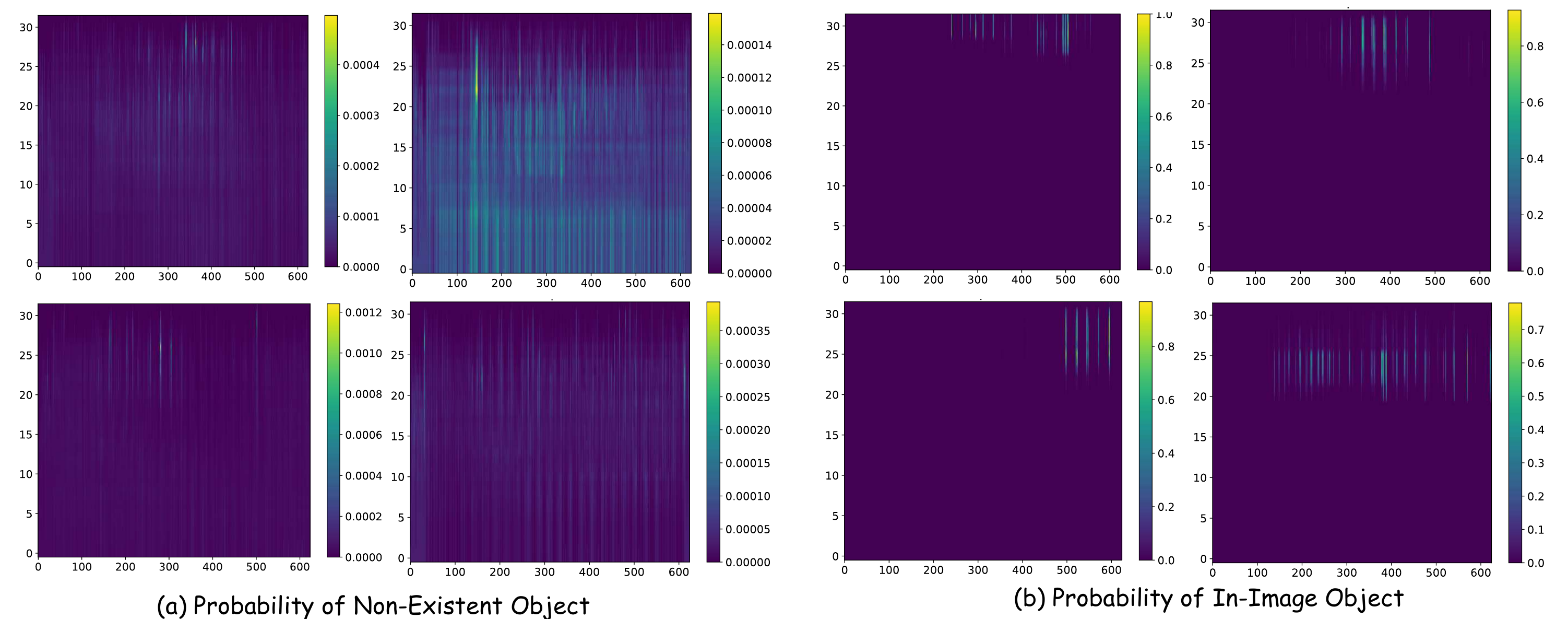}
\caption{Examples of confidence distribution for non-existent vs. in-image objects.}
\vspace{-10pt}
\label{fig:appendix-motivation}
\end{figure}

Here, we provide a more detailed analysis of our preliminary experiments described in the introduction. Specifically, we construct an analysis dataset consisting of 3,000 images sampled from a random split of the MSCOCO subset in the POPE benchmark. Each image is paired with the prompt: ``Please describe this image in detail.'' We use LLaVA-1.5 to process these image-text pairs and project all hidden states across different layers and time steps into the vocabulary distribution using the language head. Using the object annotations provided in the MSCOCO 2014 validation set, we analyze the confidence scores assigned to each object token to investigate whether the model inherently attends to factual visual entities in its internal representations. An interesting pattern is illustrated in Figure~\ref{fig:appendix-motivation}.

As shown in Figure~\ref{fig:appendix-motivation}, we observe that at least one region of the hidden states exhibits high confidence for in-image objects, while most hidden states assign very low confidence to non-existent objects. Here, tokens with high confidence for in-image objects are considered to contain real semantics related to input image, thereby possessing a certain factual information. In contrast, tokens with low confidence assigning to hallucinated objects represent non-existing semantics and provide supportive information. These results further suggest that factual information is not localized, but distributed across the model’s hidden states in different layers and time steps. Notably, we observe that factual signals tend to converge in deeper layers (i.e., layers beyond 24), which supports our decision to set a deeper refinement start layer when applying Layer-Wise Criss-Cross Attention.

\section{More Implementation Details}

\subsection{Dataset Description}
We provide descriptions of the datasets used in our experiments. POPE is designed to evaluate object-level hallucinations by checking whether a model generates non-existent visual objects. It consists of three subsets (MSCOCO, A-OKVQA, and GQA), each containing 3,000 VQA samples. MME is a comprehensive benchmark covering 14 multimodal subtasks, offering a broad evaluation of model capabilities. MMHal-Bench is an AI-assisted benchmark that evaluates open-ended responses in terms of hallucination and informativeness. For evaluation, we use the GPT-4.1 API to automatically rate model outputs.

\subsection{Implementation Details}

\begin{table}[t]
    \centering
    \begin{minipage}{0.48\textwidth}
        \centering
        \vspace{-8pt}
        \caption{Hyperparameter settings of DAMO on LLaVA and mPLUG. ``M'' denotes MME and ``P'' denotes POPE.}
        \scalebox{0.8}{ 
        \begin{tabular}{ccccc}
        \hline
                    & LLaVA M & LLaVA P & mPLUG M & mPLUG P \\
        \hline
        $\beta_{1}$ & 0.05    & 0.20    & 0.60    & 0.40      \\
        $\beta_{2}$ & 0.20    & 0.40    & 0.80    & 0.60      \\
        $\tau$      & -0.30   & -0.30   & -0.60   & -0.60  \\
        \hline
        \end{tabular}
        }
        \label{tab:damo_hyper}
    \end{minipage}
    \hfill 
    \begin{minipage}{0.48\textwidth}
        \centering
        \caption{Hyperparameters of \ours on different models and datasets. ``M'' denotes MME and ``P'' denotes POPE.}
        \scalebox{0.8}{ 
        \begin{tabular}{ccccccc}
        \hline
        \multicolumn{1}{l}{} & \multicolumn{2}{c}{LLaVA1.5} & \multicolumn{2}{c}{mPLUG-Owl2} & \multicolumn{2}{c}{InstructBLIP} \\
        \cline{2-7}
                              & M    & P    & M    & P    & M    & P    \\
        \hline
        $\ell_{\text{start}}$ & 29   & 25   & 28   & 28   & 28   & 19   \\
        $\alpha$              & 0.80 & 0.84 & 0.90 & 0.94 & 0.90 & 0.98 \\
        $\beta$                & 0.10 & 0.93 & 0.95 & 0.96 & 0.99 & 0.98 \\
        \hline
        \end{tabular}
        }
        \label{tab:ours_hyper}
    \end{minipage}
\end{table}

For LVLMs, we use three models in our experiments: LLaVA-v1.5-7B (based on Vicuna-1.5), mPLUG-Owl2 (based on LLaMA2-7B), and InstructBLIP (based on Vicuna-1.1-7B). When experimenting on the mPLUG-Owl2 and InstructBLIP, we append the prompt ``Answer the question using a single word or phrase.'' after each question on both MME and POPE. Here, the detailed hyperparameters used in \ours and other methods are listed below: 

\vspace{2pt}
\noindent{\textbf{DAMO.}}
The original DAMO paper reports hyperparameter settings 
$(\beta_1,\beta_2,\tau)$ for LLaVA and mPLUG. 
We adopt these configurations directly without additional tuning. 
For InstructBLIP, where no hyperparameters are available in prior work, 
we simply reuse the LLaVA configuration. 
The values used in our experiments are summarized in 
Table~\ref{tab:damo_hyper}.

\vspace{2pt}
\noindent{\textbf{MAP.}}
Our method involves selecting a start layer $\ell_{\text{start}}$ and two 
coefficients $(\alpha,\beta)$ for each backbone–dataset pair. 
Across different models, we observe consistent behavior: the optimal 
intervention point always lies in the late layers of the vision encoder, 
and the coefficients fall within stable ranges ($\alpha \in [0.80,0.99]$, 
$\beta \in [0.90,0.99]$), with one exception on LLaVA--MME where a smaller 
$\beta$ yields slightly better validation performance. The hyperparameters used in all experiments are summarized in 
Table~\ref{tab:ours_hyper}.

\vspace{2pt}
\noindent{\textbf{VCD.}}
For VCD, we adopt the hyperparameter settings reported in the original 
implementation and fix $\alpha=1.0$, $\beta=0.1$, and $\gamma=0.1$ for 
all backbones. Following the prior work, the noise step $T$ is set to 
999 on POPE and 500 on MME.

\vspace{2pt}
\noindent{\textbf{HGAI.}}
HGAI reports hyperparameters only for LLaVA, using 
$\text{ls}=3$, $\text{le}=13$, and $\alpha=0.5$. 
Since prior work does not provide configurations for other backbones, 
we directly reuse the LLaVA settings for mPLUG-Owl2 and InstructBLIP 
without additional tuning.

\vspace{2pt}
\noindent{\textbf{SPIN.}}
SPIN requires specifying a start layer, a routing ratio $r$, and a
blending weight $\alpha$ for each backbone: LLaVA (start-layer $=0$, $r=0.2$, $\alpha=0.1$), 
mPLUG-Owl2 (start-layer $=16$, $r=0.1$, $\alpha=0.05$), and
InstructBLIP (start-layer $=16$, $r=0.1$, $\alpha=0.08$). 
We observe that the performance is relatively stable under small
perturbations around these values, and therefore we fix them for all
datasets for each backbone.

\vspace{2pt}
\noindent{\textbf{DCLA.}}
DCLA provides a single set of hyperparameters $(\alpha,\tau)$ for each 
backbone, and these settings are stable across benchmarks. 
For LLaVA, we adopt the values reported in the original paper 
($\alpha=0.82$, $\tau=0.74$).  
For mPLUG-Owl2, we use the corresponding configuration 
($\alpha=0.90$, $\tau=0.95$) provided by the original implementation. 
Since prior work does not report DCLA settings for InstructBLIP, 
we directly reuse the LLaVA configuration.

\begin{table}[t]
\centering
\footnotesize
\setlength{\tabcolsep}{3.5pt} 
\renewcommand{\arraystretch}{1.15}
\caption{Detailed ablation results of various map-level operations across all MME subtasks. All experiments are conducted on LLaVA-1.5-7B.}
\resizebox{\textwidth}{!}{%
\begin{tabular}{l c c c c c c c c c c c} 
\toprule
Map Operation & Cel. & Col. & Art. & Cou. & Lan. & Exi. & Pos. & OCR & Psi. & Sce. & Total \\
\midrule
Vanilla & 135.00 & 165.00 & \textbf{119.25} & 160.00 & 162.25 & 190.00 & 140.48 & 125.00 & 138.33 & 156.25 & 1491.56 \\
Global Attn. & 135.29 & 170.00 & 118.00 & 156.67 & 161.25 & 190.00 & 137.41 & 140.00 & 143.33 & 157.75 & 1509.71 \\
Local (7$\times$7) & 133.53 & 165.00 & 118.25 & 156.67 & 160.50 & 190.00 & 137.41 & 140.00 & 143.33 & 157.75 & 1502.44 \\
Local (5$\times$5) & 135.29 & 165.00 & 118.00 & 156.67 & 159.75 & 190.00 & 137.41 & 140.00 & 143.33 & 157.00 & 1502.46 \\
Criss-Cross & 133.53 & 170.00 & 117.25 & 161.67 & 160.50 & 190.00 & 136.39 & 132.50 & \textbf{148.33} & 157.75 & 1507.92 \\
Layer-Wise Criss-Cross & \textbf{137.94} & 165.00 & 115.00 & \textbf{165.00} & \textbf{165.25} & \textbf{190.00} & \textbf{144.56} & \textbf{140.00} & 138.33 & \textbf{159.25} & \textbf{1520.33} \\
\bottomrule
\end{tabular}}
\label{tab:appendix-map-ablation}
\end{table}

\section{Additional Results for Ablation Studies}

\subsection{Detailed Results across Map-Level Operation}
Table~\ref{tab:appendix-map-ablation} presents the full MME subtask 
results corresponding to the ablation study in the main paper. 
Overall, Layer-Wise Criss-Cross Attention achieves 
the strongest performance across most categories, confirming the 
effectiveness of progressively aggregating map-level information along 
the depth of the vision encoder. The variant with broadcast further 
improves the overall MME score.

Across subtasks, we observe several consistent patterns. All map-level 
operations substantially enhance OCR-related performance compared to the 
vanilla baseline, indicating improved fine-grained factual grounding. 
Spatially oriented subtasks such as \textit{Position} and 
\textit{Scene} also benefit from stronger structural fusion mechanisms: 
global attention and criss-cross attention perform well, while the 
layer-wise variant yields the highest scores in both cases. 
On the other hand, the \textit{Artwork} subtask shows a slight drop 
across all variants, suggesting that this category may be less sensitive 
to enhanced token-level structural aggregation.

\begin{table}[t]
\centering
\footnotesize
\setlength{\tabcolsep}{3.5pt} 
\renewcommand{\arraystretch}{1.15}
\caption{Detailed MME scores on LLaVA-1.5-7B across different $\beta$ values. All runs share the same $\ell_{\text{start}}$ and $\alpha$ settings.}
\resizebox{\textwidth}{!}{%
\begin{tabular}{l c c c c c c c c c c c}
\toprule
$\beta$ & Cel. & Col. & Art. & Cou. & Lan. & Exi. & Pos. & OCR & Psi. & Sce. & Total \\
\midrule
vanilla & 135.00 & 165.00 & \textbf{119.25} & 160.00 & 162.25 & 190.00 & 140.48 & 125.00 & 138.33 & 156.25 & 1491.56 \\
0.90    & 138.24 & 165.00 & 115.00        & 165.00 & 165.25 & 190.00 & 144.56 & 140.00 & 138.33 & 159.25 & 1520.63 \\
0.91    & 137.94 & 165.00 & 114.25        & 160.00 & 164.50 & 190.00 & 141.50 & 140.00 & 138.33 & 159.25 & 1510.77 \\
0.92    & 137.94 & 165.00 & 115.00        & 160.00 & 164.50 & 190.00 & 141.50 & 140.00 & 138.33 & 159.25 & 1511.52 \\
0.93    & 138.24 & \textbf{170.00} & 115.75 & \textbf{165.00} & 165.25 & \textbf{190.00} & 142.52 & \textbf{140.00} & \textbf{143.33} & \textbf{159.25} & \textbf{1529.34} \\
0.94    & \textbf{139.12} & 170.00 & 115.75 & 160.00 & 165.25 & 190.00 & 143.54 & 140.00 & 143.33 & 159.25 & 1526.24 \\
0.95    & 137.94 & 165.00 & 114.25 & 165.00 & \textbf{165.25} & 190.00 & \textbf{144.56} & 140.00 & 138.33 & 159.25 & 1519.58 \\
\bottomrule
\end{tabular}}
\label{tab:appendix-beta-ablation}
\end{table}

\subsection{Detailed Results for $\beta$ Ablation}
Table~\ref{tab:appendix-beta-ablation} presents the per-category MME
scores on LLaVA1.5 for different values of $\beta$ in the range
$[0.90, 0.95]$, corresponding to the ablation study in the main paper.
Across all settings, \ours consistently outperforms the vanilla baseline
(1491.56 total), indicating that the method is stable under moderate
variations of $\beta$. It can be observed that most categories, such as \textit{OCR}, \textit{Landmark} and \textit{Posters}, exhibit minimal sensitivity to
$\beta$, with all values yielding strong and nearly identical scores. Overall, $\beta=0.93$ provides the best trade-off across
categories, achieving the highest MME total value of 1529.34.

\subsection{More Qualitative Results}
\begin{figure}[t]
\centering
\includegraphics[width=1.0\columnwidth]{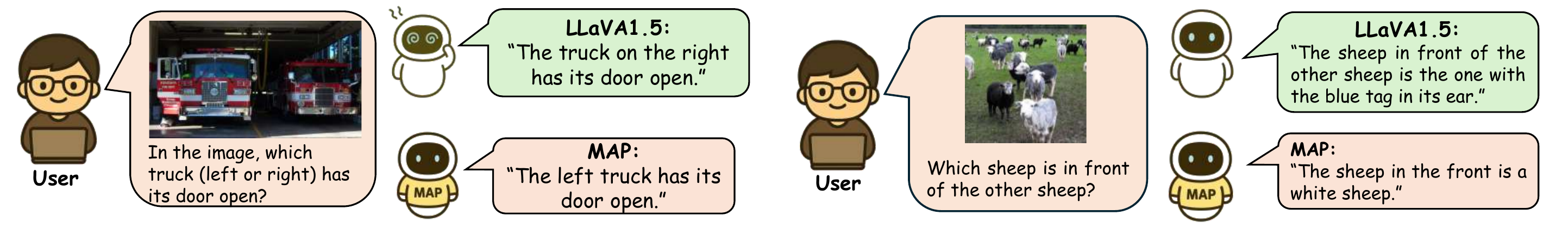}
\caption{Comparison of LLaVA1.5 and \ours with qualitative examples in MMHal-Bench.}
\label{fig:appendix-MMhal}
\end{figure}

MMHal-Bench evaluates open-ended responses from the perspectives of 
both informativeness and hallucinations. Figure~\ref{fig:appendix-MMhal} shows two additional qualitative examples comparing \ours with 
LLaVA-1.5. In Figure~\ref{fig:appendix-MMhal} left, the model is asked which truck has its door open. LLaVA-1.5 hallucinates the wrong side, while \ours correctly identifies the truck on the left, rectifying the existence error. In Figure~\ref{fig:appendix-MMhal} right, LLaVA-1.5 produces an irrelevant description about a sheep with a blue tag, whereas \ours gives a concise but faithful description grounded in the foreground sheep, avoiding additional hallucinated details and yielding 
a more reliable response.

\end{document}